\documentclass{article}

\usepackage{arxiv}

\usepackage[utf8]{inputenc} 
\usepackage[T1]{fontenc}    
\usepackage{hyperref}       
\usepackage{url}            
\usepackage{booktabs}       
\usepackage{amsfonts}       
\usepackage{nicefrac}       
\usepackage{microtype}      
\usepackage{lipsum}

\usepackage{amsmath}
\usepackage{multirow}
\usepackage[]{algorithm2e}
\usepackage{subfigure}
\usepackage{epsfig}
\usepackage{timesmt}

\DeclareMathOperator{\argmin}{arg\,min}

\title{The multi-objective optimisation of breakwaters using evolutionary approach}

\author{
  Nikolay O. Nikitin \\
  National Center for Cognitive Technologies\\
  ITMO University\\
  49 Kronverksky Pr., St. Petersburg, 197101, Russia\\
  \texttt{nnikitin@itmo.ru} \\
   \And
 Iana S. Polonskaia \\
 National Center for Cognitive Technologies\\
  ITMO University\\
  49 Kronverksky Pr., St. Petersburg, 197101, Russia\\
    \texttt{ispolonskaia@itmo.ru} \\
\And
 Anna V. Kalyuzhnaya\\
 National Center for Cognitive Technologies\\
  ITMO University\\
  49 Kronverksky Pr., St. Petersburg, 197101, Russia\\
 \And Alexander V. Boukhanovsky\\
 National Center for Cognitive Technologies\\
  ITMO University\\
  49 Kronverksky Pr., St. Petersburg, 197101, Russia\\
}

\begin{document}
\maketitle

\begin{abstract}
In engineering practice, it is often necessary to increase the effectiveness of existing protective constructions for ports and coasts (i. e. breakwaters) by extending their configuration, because existing configurations don't provide the appropriate environmental conditions. That extension task can be considered as an optimisation problem. In the paper, the multi-objective evolutionary approach for the breakwaters optimisation is proposed. Also, a greedy heuristic is implemented and included to algorithm, that allows achieving the appropriate solution faster. The task of the identification of the attached breakwaters optimal variant that provides the safe ship parking and manoeuvring in large Black Sea Port of Sochi has been used as a case study. The results of the experiments demonstrated the possibility to apply the proposed multi-objective evolutionary approach in real-world engineering problems. It allows identifying the Pareto-optimal set of the possible configuration, which can be analysed by decision makers and used for final construction.
\end{abstract}


\section{INTRODUCTION}
\label{sec:intro}
To identify the optimal breakwater layout for a specific problem, the decision maker (that usually is a strong expert in the field) should propose or choose the configuration which best matches the set of cost and quality criteria \cite{steuer1986multiple}. In most cases, the decision maker tries to find the solution based on previous experience and the subject area knowledge. This problem is often considered as a minimisation of the risks \cite{sorensen2004risk}, \cite{alises2014overtopping} during the coastal structures life cycle.

However, the optimal design problem is difficult for expert analysis, especially in the case of high-dimensional problems, because the expert should take into account all objective functions, problem constraints and check a large number of new construction configurations to find a near-optimal solution. Therefore, the intelligent optimisation methods (e.g. evolutionary algorithms) are widely used for this task \cite{lagaros2002structural}. Also, the involvement of the numerical metocean models make it possible to simulate the local environmental conditions for different layouts of the protective constructions and obtain the values of quality metrics for proposed configuration.

The existing approaches to this task and similar structural optimisation problems are considered in Section \ref{sec:related}. In Section \ref{sec:problem}, the formulation of the breakwater optimisation task as a multi-objective optimisation problem is described. It includes several objective functions and constraints. Also, this section includes the description of the SWAN wind-wave model's configuration for the port water area that is used for wave height estimation. Several implemented evolutionary approaches are presented in Section \ref{sec:evo}. Section \ref{sec:exp} includes the description of the experiments' setup and the experimental results for the breakwaters layout optimisation in the Sea Port of Sochi as a case study. Finally, conclusions and future plans are presented in Section \ref{sec:conclusions}.
\section{RELATED WORK}
\label{sec:related}
Each particular optimal design problem has a lot of peculiarities, and it is necessary to take them into account during the algorithm developing process. However, it is useful to create some generalized approach which could be applied for a broad range of breakwaters' optimisation problems and could provide the opportunity to select the optimisation objectives and constraints for different real-world cases. The correct selection of objective function and constraints is the main part of the considered optimisation problem-solving process. There are a lot of objectives and constraints that can be involved. The comprehensive survey of constraints and criteria that can be taken into attention in problems of optimal harbor's design identification and improvement is provided in \cite{diab2017survey}.

Usually, authors simplify the analysed problem to find a solution faster, because time and computational resources are strictly limited in many cases. Thus, in \cite{elchahal2013optimization}, a genetic optimisation approach for the design of small detached breakwaters with three segments is used. It considers wave disturbance inside the port and ship manoeuvring constraints. In \cite{diab2014optimisation} the two-segment breakwater was optimised using an evolutionary algorithm in the strictly restricted range of spatial coordinates, directly transformed to a numerical chromosome. A wave disturbance constraint was added to the fitness function. Also, two other constraints related to breakwater geometry characteristics have been taken into account. Decisions with unsatisfactory geometry characteristics (incorrect angle between two segments or undesired location) are discarded during optimisation. In both papers, each segment of the breakwater is presented in the chromosome using the end coordinates.

Besides the spatial coordinates of breakwater's segments, there are other breakwater characteristics that can be optimised. For example, in \cite{nikoo2014multi} and \cite{somervell2017novel}, the structural parameters of the double-layered breakwater were optimised using multi-objective genetic algorithms. In the first case, three parameters of a compound breakwater were optimised: porosities of each breakwater wall and the gap between them, in the second case five parameters such as porosities of the upper part of the wall for each breakwater layer, the relative depths of submergence of upper parts of these walls into the water and the relative gap between the vertical walls. Optimisation problem statement combines two conflicting objectives: maximisation of the wave energy dissipation coefficient and minimisation of the material volume in the breakwater in the first case. The second case also combined conflicting wave reflection and wave transmission objectives.

It is noticed that a lot of marine optimisation engineering problems can be solved more reliably using multi-objective optimisation approaches. Thus, in \cite{elkinton2008algorithms}, the authors compared five different optimisation approaches, including evolutionary algorithms, to optimise the layout of an offshore windfarm. A fitness function is based on a trade-off between maximum energy production on the farm and minimum cost. A similar problem was solved in \cite{gonzalez2019surrogate} for the array of tidal turbine generators. To reduce the computational cost of simulations, the surrogate model was used during optimisation. The optimisation of fairway design parameters was considered in \cite{gucma2019optimization}. In \cite{rustell2014decision}, a genetic algorithm was used for the selection of optimal structural parameters for the gas terminal layout. The genetic chromosome representation is based on nine different characteristics describing the certain terminal configuration. The fitness function includes three objectives (capital cost, maintenance cost and vessel downtime due to waves) which are simultaneously minimised. The proposed algorithm allows finding a few configurations that are better than the design obtained using the traditional approach and decreases the cost by a factor of 30.

However, the described approaches are tested mostly for specific simplified cases. The proposed algorithms are aimed to resolve low-dimension tasks with a short chromosome, that can describe the real harbour constructions only in a coarse way. For that reason, it is useful to develop the multi-objective approach for the breakwater optimisation and validate it with a real-world test case.

\section{PROBLEM STATEMENT}
\label{sec:problem}
\subsection{Breakwater design as optimisation task}

\begin{figure}
\centerline{\includegraphics[width=14cm]{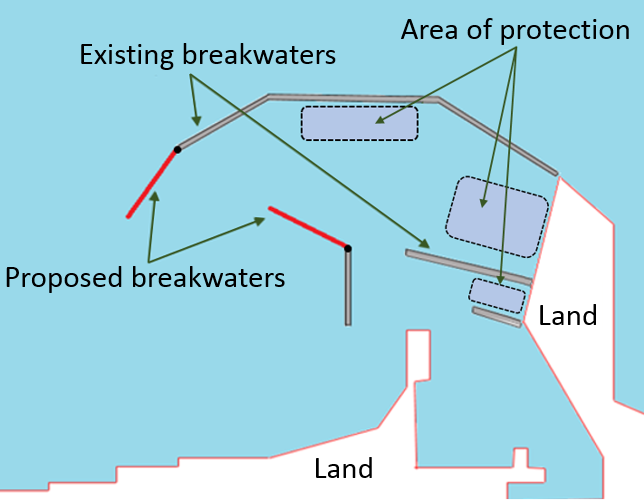}}
\caption{The scheme of harbor with existing breakwaters and the structure of attached breakwaters that should be optimised}
\label{f:common_harbor_for_opt}
\end{figure}

The extension of existing breakwaters' configuration can be separated into several aspects: the selection of the attached segments number, the attachment points' location for the new segments, the spatial area for defence and coordinates of control points. These variables can be involved in the optimisation problem, but they can also be defined in the preliminary step by the economic and engineering criteria and then considered as static. The optimal design task for this case is presented in figure~\ref{f:common_harbor_for_opt}.

The problem  of breakwater design identification can be formulated as multi-objective function optimisation in the space of structural parameters of breakwaters and written as:
\begin{equation} \begin{split}
\label{eq_opt}
& \theta_{opt} = \argmin_{\theta}{F(\theta)}, \\
& F(\theta) = \mathcal{G}(H_j(Y(\theta)), C(\theta), N(\theta)), \\
\end{split} \end{equation}
where $\mathcal{G}(\bullet)$ is an operator for multiobjective transformation to a function $F$ for a given structural parameters set $\theta$, $H_j$ is the wave height objective function for certain control points, $C$ is the objective function for construction cost, N is objective function for manoeuvring safety, $Y$ is the environmental characteristics simulation results. 

There are additional objectives exists that can be taken into account during optimisation using the proposed approach (the implemented software architecture allow to specify the custom objective functions set). For example, the resonant periods associated with eigenfrequencies in the port area can be calculated to estimate the possibility of harbor seiches' occurrence. However, the experimental studies in this paper are based on the objectives described in subsection \ref{subsec:objectives}.

\subsection{Structural parameterisation}
There are various parameterisation approaches that can be used to represent the optimisable structure in the most efficient way \cite{samareh2001survey}. The breakwaters' layout can be represented as a grid with binary (solid/wet) values in each cell; as a set of absolute or relative spatial (Cartesian) coordinates; as parameters of approximating function (polynomial or spline). The most widely used approach for breakwater structure representation - Cartesian encoding - has several disadvantages for numerical optimisation. This encoding does not allow to modify the length and direction of the breakwater segment separately \cite{bendsoe1988generating}. So, it is not convenient to "rotate" the part of a breakwater to an optimal angle using crossover or mutation operators of the genetic algorithm. 

\begin{figure}
\centerline{\includegraphics[width=14cm]{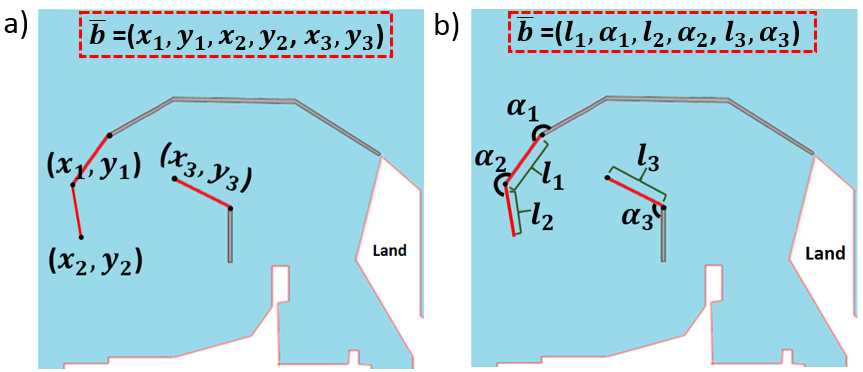}}
\caption{The breakwater parameterisation in different encoding: a) with Cartesian coordinates b) with angular coordinates}
\label{f:genotype_encodings}
\end{figure}

To resolve this issue, the modified approach based on encoding in relative polar coordinates is proposed. Since the transformation of Cartesian coordinates to polar ones is objective, there is no structural simplification in this step (the relative position of each segment is still represented by two numerical values - angle and direction instead of X and Y values for Cartesian coordinates). However, the evolutionary transformations of a polar chromosome can be implemented in a more effective way. The Cartesian and angular chromosomes are represented in figure~\ref{f:genotype_encodings}.

\subsection{Objective functions}
\label{subsec:objectives}
The correct selection of the objective functions is an important part of the formalisation of the real-world breakwater design issue to the optimisation problem. As it noted in \cite{diab2017survey}, there are a lot of various criteria (environmental, hydro-dynamical, mechanical, economical, manoeuvring, etc.) that should be taken into attention during the  decision-making process. However, to construct the objective function, it is necessary to implement the algorithm for their numerical evaluation. To evaluate the metric for the navigational objective, the complex port simulation model \cite{olba2018state} or the ship path model \cite{sutulo2002mathematical} can be used. Cost objectives can be based on probabilistic economic models \cite{piccoli2014economic}. Also, the additional penalty for the complexity of solution (e.g. number of segments with non-zero length) can be involved, but usually, the cost objective takes it into account in a non-direct way)

In a frame of proposed multi-objective evolutionary algorithms, several objectives are used. They are described in table~\ref{t:obj}. In addition to the main objectives (cost, navigational, and wave height-based), structural constraints were added. They restrict the generation of unrealistic solutions (self-crossing and land-covering breakwaters) during the evolutionary optimisation because such configurations can cause unstable behaviour of the wind-wave model.

\begin{table}[]
\small
\caption{\label{t:obj}The objective functions and constraints considered in breakwater optimisation problems}
\begin{center}
\begin{tabular}{|l|l|l|c|}
\hline
ID & Name& Description & \multicolumn{1}{l|}{\begin{tabular}[c]{@{}l@{}}Measurement\\ unit\end{tabular}}                                                 \\ \hline
1 & \begin{tabular}[c]{@{}l@{}}Relative cost\\ objective\end{tabular} & \begin{tabular}[c]{@{}l@{}}Total cost of new\\ breakwaters\\ construction\end{tabular}& \multirow{5}{*}{\begin{tabular}[c]{@{}c@{}}\% of increase\\ against base\\ configuration\\ (the less is\\ better)\end{tabular}} \\ \cline{1-3}
2 & \begin{tabular}[c]{@{}l@{}}Relative\\ navigational\\ objective\end{tabular} & \begin{tabular}[c]{@{}l@{}}The safety of ships’\\ maneuvering\\ scheme represented\\ as minimum length\\ between each point\\ of new breakwaters\\ and central line of\\ fairways\end{tabular} &      \\ \cline{1-3}
3 & \multirow{3}{*}{\begin{tabular}[c]{@{}l@{}}Relative wave\\ height\\ objective\end{tabular}} & \multirow{3}{*}{\begin{tabular}[c]{@{}l@{}}Wave height in\\ each control point\end{tabular}} & \\ \cline{1-1}
4 & &  & \\ \cline{1-1}
5 & & & \\ \hline
6  & \begin{tabular}[c]{@{}l@{}}Structural\\ constraint\end{tabular} & \begin{tabular}[c]{@{}l@{}}Self-intersections\\ are not allowed\end{tabular} & \multicolumn{1}{l|}{\begin{tabular}[c]{@{}l@{}}Number of self-\\ intersections\end{tabular}} \\ \hline
7  & \begin{tabular}[c]{@{}l@{}}Navigational\\ constraint\end{tabular} & \begin{tabular}[c]{@{}l@{}}Breakwater\\ shouldn't intersect\\ fairway center line\end{tabular} & \multicolumn{1}{l|}{\begin{tabular}[c]{@{}l@{}}Number of\\ fairway center\\ line\\ intersections\end{tabular}} \\ \hline
8  & \begin{tabular}[c]{@{}l@{}}Relative \\ quality \\ objective\end{tabular} & \begin{tabular}[c]{@{}l@{}}Used in single\\- objective \\ evolutionary \\ algorithm and \\ should consider \\ all necessary\\ objectives\end{tabular} & \multicolumn{1}{l|}{\begin{tabular}[c]{@{}l@{}}The composition \\ of the \\ rest relative\\  objectives\end{tabular}} \\ \hline
\end{tabular}
\end{center}
\end{table}

The value of cost objective is directly proportional to summary breakwaters' length:
\begin{equation} \begin{split}
\label{cost_obj}
&C=\sum_{i=1}^{n}\sqrt{\left (x_{2,i}-x_{1,i}\right )^{2}+\left (y_{2,i}-y_{1,i}\right )^{2}}\cdot s,\\ 
\end{split}
\end{equation}
where $n$ is a number of breakwaters' segments, $s$ is a grid step.

The value of the wave height objective is a vector of significant wave heights $h$ in control points (where $m$ is the number of control points):
\begin{equation} \begin{split}
\label{wh_obj}
&H=(h_{1},h_{2},...,h_{m}) 
\end{split} \end{equation}

\begin{figure}
\centerline{\includegraphics[width=12cm]{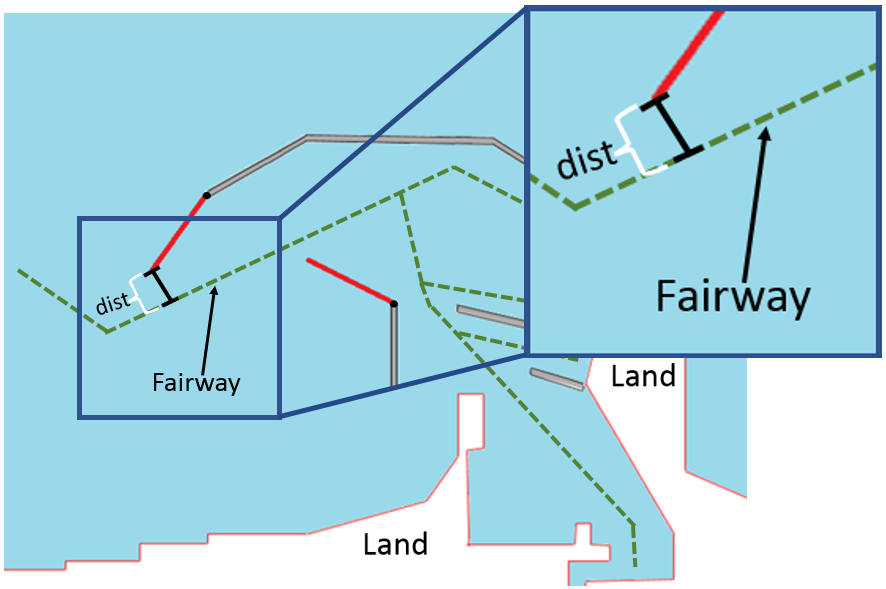}}
\caption{The concept of navigational objective based at minimal fairway distance}
\label{f:nav_obj}
\end{figure}

The navigational objectives are evaluated by a simple geometrical model described in figure~\ref{f:nav_obj} and written as:
\begin{equation} \begin{split}
\label{nav_obj}
&N=min(||p_{i,j}^{b}-p_{k}^{f}||),\\
&i=\overline{1,n_{1}}, j=\overline{1,n_{2}}, k=\overline{1,n_{3}}
\end{split} \end{equation} 
where $n_{1}$ is a number of breakwaters, $n_{2}$ is a number of breakwater points, $n_{3}$ - number of fairway points, and $p^{b}$ and $p^{f}$ are breakwater and fairway points.

Relative value of each objective type is calculated as:
\begin{equation} \begin{split}
\label{rel_obj}
f_{i}^{rel}=\frac{f_{i}^{new}-f_{i}^{old}}{f_{i}^{old}}\cdot 100\%, i=\overline{1,t} 
\end{split} \end{equation}
where $t$ is a number of objectives.

The objective function for the single-objective optimization approach is presented as the following convolution of objectives:
\begin{equation} \begin{split}
\label{rel_qual_obj}
f_{single}^{rel}=\frac{100+mean(f_{2}^{rel})+f_{3}^{rel}}{100-f_{1}^{rel}},
\end{split} \end{equation}

This function combines all objectives to achieve the best relative quality of obtained solutions.

\subsection{Wave simulation}
\label{sec:wavesim}

The analysis of the effectiveness of a specific breakwater layout requires the corresponding numerical simulation of met-ocean conditions for the proposed harbour design. The numerical models of wind-wave interaction allow calculating the wave characteristics based on depth, wind direction and constructions configuration in the harbour. A common approach is to use the third-generation wind-wave model for this purpose. According to the community experience \cite{niculescu2018evaluation}, the spectral wind-wave model SWAN \cite{booij1999third} was used to reproduce the integral wave characteristics in the desired area. Numerical simulation was performed on a regular grid with cell size 25x25m. The parameters of extreme storms were used as a boundary conditions for simulation. The spatial structure of breakwaters' configurations was represented using obstacles (the reflection coefficients was set differently for the solid wall and tetrapod breakwaters).

 The simulations were run for each breakwaters' configuration generated by the evolutionary algorithm. The estimation of the wave height was carried out in several control points. 

The simultaneous reproduction of the reflection and diffraction effects inside harbor by the SWAN model is quite limited \cite{violante2009diffraction}. However, the better simulation of the diffraction effect can be achieved by modification of the model as described in \cite{kim2017improving}.

\begin{figure}
\centerline{\includegraphics[width=12cm]{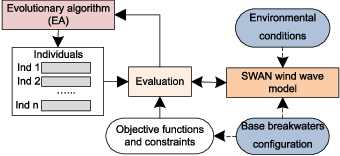}}
\caption{The scheme of interaction between of evolutionary algorithm and wind wave model during optimisation}
\label{f:evo_interaction}
\end{figure}

The scheme of interaction between the evolutionary algorithm and the SWAN model in the frame of the described optimisation task is presented in figure~\ref{f:evo_interaction}.

\section{EVOLUTIONARY APPROACHES TO THE BREAKWATER OPTIMISATION}
\label{sec:evo}

\subsection{Multi- and single- objective evolutionary optimisation}
Evolutionary algorithms (EA) are population-based methods for global search that especially demonstrate their advantages over classical optimization approaches in multi-objective and multi-dimension optimization problems. In addition, this class of algorithms can easily be adapted to the problem at hand. In this study, the multi-objective approach - the Strength Pareto Evolutionary Algorithm (SPEA2) and the single-objective evolutionary approach - Differential evolution (DE) are used for solving the multi-objective breakwaters optimization problem. 

The selection of parents is the first stage of evolutionary optimisation, then, the crossover and mutation of selected individuals are performed. The main difference between using multi-objective and single-objective genetic algorithms consists in the parents' selection strategy and the fitness evaluation strategy. Tournament selection is used in single-objective DE and the fitness function is defined as the combination of all optimisation criteria (objectives described in Section \ref{sec:problem}). The multi-objective approach uses the special type of selection adapted to multi-criteria problems and includes few steps (environmental selection and binary tournament selection).

\begin{figure}
\centerline{\includegraphics[width=14cm]{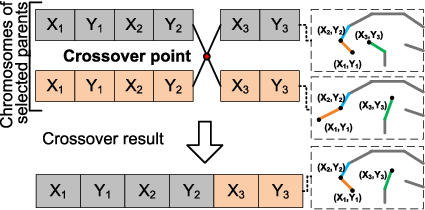}}
\caption{Illustration of the one-point crossover for 3-segment breakwater}
\label{f:onepoint_crossover}
\end{figure}

The crossover operator allows the individuals to interchange the genetic material. Thus the algorithm has an opportunity to combine effective breakwater's segments from different individuals. Figure~\ref{f:onepoint_crossover} illustrates the crossover operator which is applied for the breakwater design problem with three attached segments. According to this concept, two selected parents swap genetic information between blocks created by chromosome separation at a randomly generated crossover point. In this implementation (pairwise crossover), the spatial coordinates (Cartesian or angular) belonging to the same breakwater's segment cannot be separated by a crossover point. 

\begin{figure}
\centerline{\includegraphics[width=14cm]{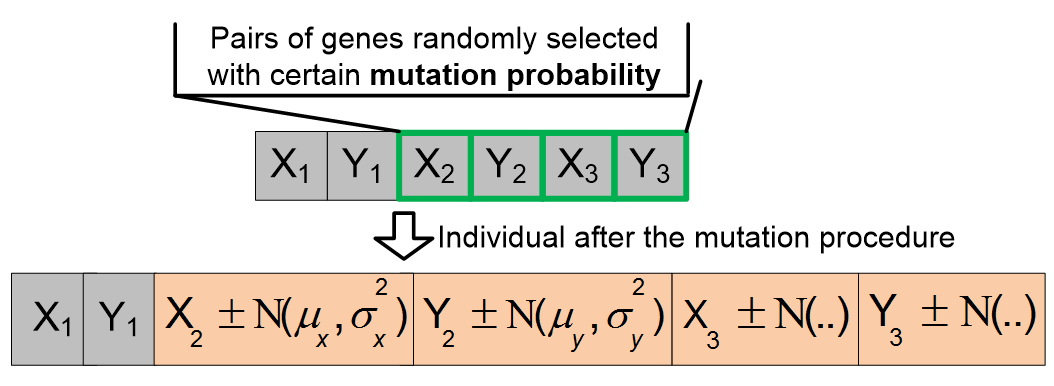}}
\caption{Mutation process representation for N-segment breakwater}
\label{f:mutation}
\end{figure}

Mutation introduces random changes to chromosome genes that allow the algorithm to avoid local minima. The implemented mutation operator for breakwaters' chromosome is demonstrated in figure~\ref{f:mutation}. First, genes are selected with a certain mutation probability, then the randomly generated values obtained from normal distribution $N(\mu,\sigma^2)$ ($\mu$ and $\sigma$ are set for each variable type) are added to genes that were selected on the previous stage. 

\subsection{Greedy evolutionary heuristics}
\begin{figure}
\centerline{\includegraphics[width=14cm]{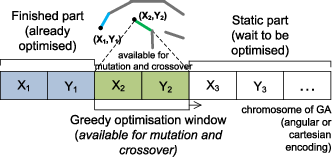}}
\caption{The chromosome modification with greedy approach to evolutionary optimisation}
\label{f:greedy_approach}
\end{figure}

The greedy algorithms perform a locally optimal choice at each optimisation stage in order to obtain global or local optimum in the end \cite{devore1996some}. Greedy heuristics in evolutionary algorithms can be used to intensify the convergence and improve the quality of the obtained solutions \cite{julstrom2005greedy}. The idea of a greedy evolutionary application for the breakwater optimisation is to separate the complex task of simultaneous optimisation of all breakwaters' segments to several stages of individual optimisations for each segment. The application of this concept for the breakwaters' optimal design task is presented in figure~\ref{f:greedy_approach}.

The pseudocode of the final implementation of the greedy heuristic in a frame of the SPEA2 evolutionary algorithm is presented in Alg.~\ref{alg_rspea2}.

\begin{algorithm}
 \KwData{populationSize, \\ archiveSize, crossoverRate, mutationRate}
 \KwResult{best individual from archive}
 \SetKwData{Archive}{archive}
 \SetKwData{Pop}{pop}
  \SetKwData{Mask}{greedyMask}
 \SetKwData{ShiftRight}{ShiftRight}
 \SetKwData{Union}{union}
 \SetKwData{Ind}{individ}
 \SetKwData{Model}{waveModel}
 \SetKwData{Objs}{waveHeightObj, costObj, navigationalObj}
 \SetKwData{Mates}{matingPool}
 \SetKwFunction{InitPop}{InitPopulation}
 \SetKwFunction{Converged}{ConvergenceCriterion}
 \SetKwFunction{Objectives}{CalculateObj}
 \SetKwFunction{RunSimulation}{RunSimulation}
 \SetKwFunction{Mean}{Mean}
 \SetKwFunction{Fit}{CalculateFitness}
 \SetKwFunction{Pareto}{TakeNonDominated}
 \SetKwFunction{Select}{BinaryTournamentSelection}
 \SetKwFunction{Variation}{CrossoverAndMutation}
 \Pop $\leftarrow$ \InitPop{populationSize} \\
 \Archive $\leftarrow \emptyset$ \\
 \While {not ConvergenceCriterion()}  {
    \For {\Ind in \Pop} {
        \Objs $\leftarrow$ \Objectives{\Ind, \Model}
     }
    \Union $\leftarrow$ \Archive + \Pop \\
    \For {\Ind in \Union} {
        \Ind.fitness $\leftarrow$ \Fit{\Ind}
    }
    \Archive $\leftarrow$ \Pareto{\Union, archiveSize} \\
    \Mates $\leftarrow$ \Select{\Union, populationSize} \\
    \Pop $\leftarrow$ \Variation{\Mates, greedyMask, crossoverRate, mutationRate}
    \Mask $\leftarrow$ \ShiftRight{\Mask}
 }

 \caption{The pseudocode of the implemented greedy multi-objective algorithm based at SPEA2 algorithm}
 \label{alg_rspea2}
\end{algorithm}

\section{EXPERIMENTAL STUDIES}
\label{sec:exp}

\subsection{Case study}

\begin{figure}
\centerline{\includegraphics[width=14cm]{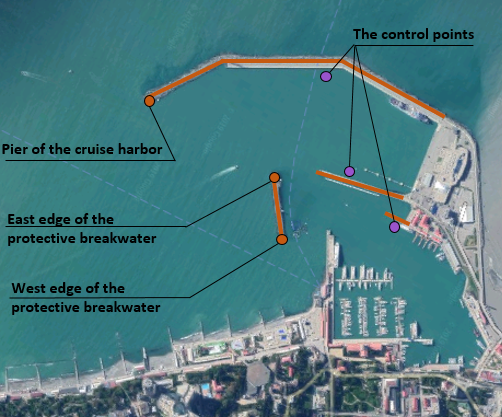}}
\caption{The satellite image of Sea Port of Sochi. The orange points represent locations for new attached breakwaters, the orange lines represents existing constructions}
\label{f:sochi_harbor}
\end{figure}

As far as it is necessary to validate the proposed evolutionary approach in the real-world case, the harbour of Sea Port of Sochi was selected as an experimental case study. The port is located in 43.58N, 39.71E (the coast of the Black Sea). The configuration of harbour and breakwaters are presented in figure~\ref{f:sochi_harbor}. As can be seen, there are two existing protection structures: the main breakwater attached to the coast and the additional detached breakwater.

The actual problem for this port is the events of strong pitching of moored ships caused by rare extreme weather conditions (storms). The proposed solution is to build additional breakwaters in addition to the existing one. It raises the task of optimal design identification. 

To use this task as a case study, we simplify it to make the experimental results of the algorithms validation more clear and interpretable. There are three static points for breakwaters' attachment. Each additional breakwater consists of two segments.

\subsection{Experimental results}

The plan of experiments consists of several stages:
1. Experimental validation of SOEA with angular and Cartesian encoding for the attached breakwaters optimisation problem;
2. Experimental validation of MOEA with both encoding for the same problem;
3. Estimation of a greedy heuristic for both algorithms.
4. Analysis of stability for obtained solutions.

\begin{figure}
\centerline{\includegraphics[width=10cm]{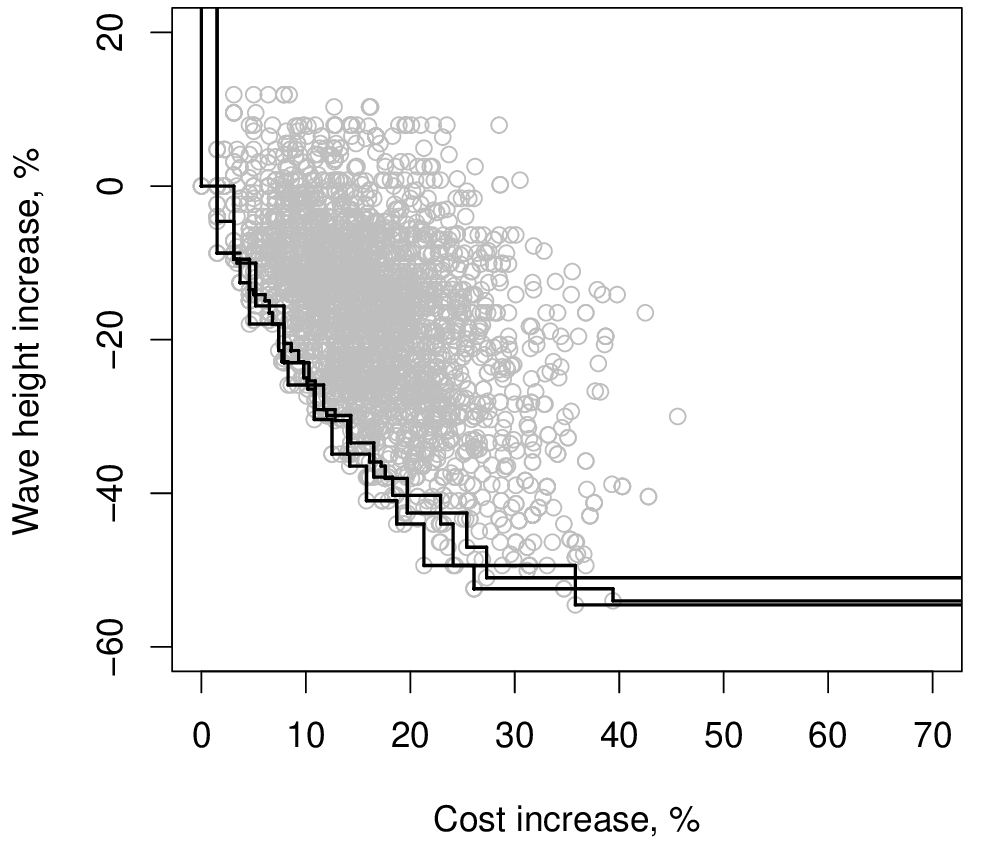}}
\caption{The example of variability of Pareto fronts (non-dominated points) for five independent optimisation runs. The gray color represents the dominated-points}
\label{f:pareto_var}
\end{figure}

During the optimisation, 30 generations with 30 individuals in each generation were examined. The figure~\ref{f:pareto_var} represents the example of Pareto fronts variability (for the multi-objective algorithm) for five independent runs. The values of the cost-based objective function and the averaged wave height objective function in different points were used to obtain 2-dimensional fronts for visualisation. It can be seen that the common pattern is the same, but not all non-dominating solutions are found during each run.

\begin{figure}
\centerline{\includegraphics[width=10cm]{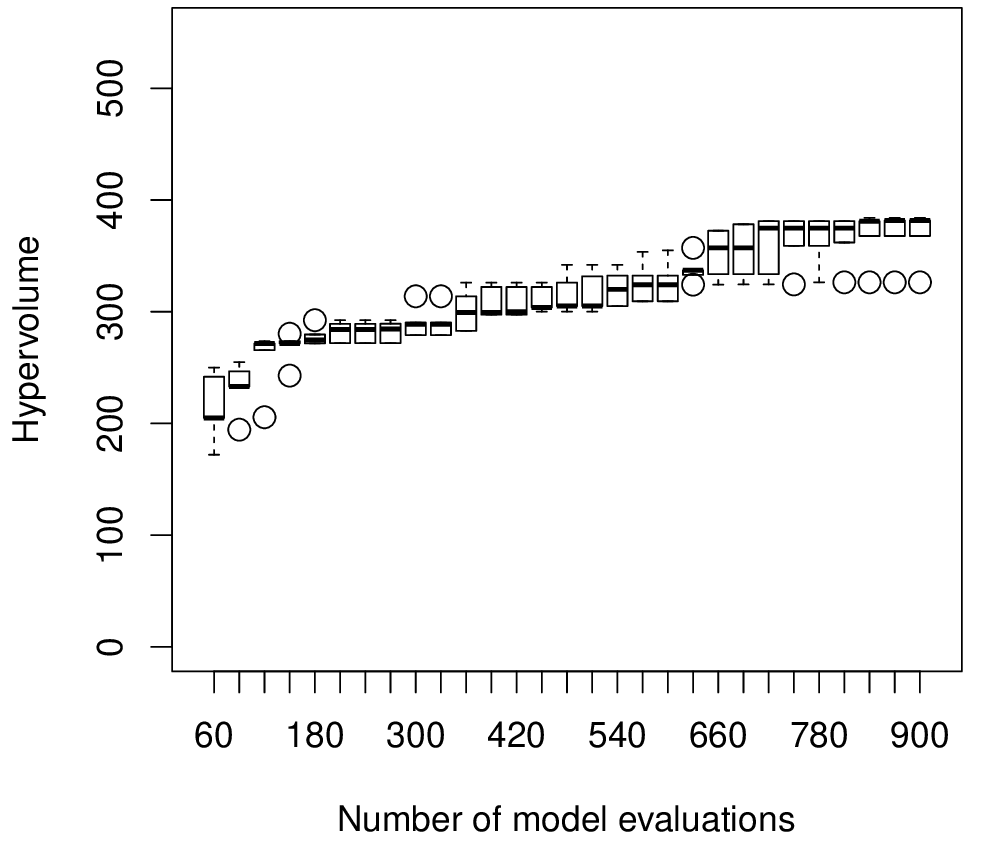}}
\caption{The boxplots of convergence of hypervolume metric in objectives space for five independent optimisation runs}
\label{f:hypervol_var}
\end{figure}

The hypervolume indicator is widely applied in the evaluation of MOEA effectiveness \cite{chugh2019multiobjective} and analysis of convergence. The figure~\ref{f:hypervol_var} illustrates the variability of non-dominated hypervolume for five independent optimisation runs. Despite the fact that one of the runs converges not in the best-found value of the hypervolume, other runs achieved convergence in the same range.

\begin{figure}
\centerline{\includegraphics[width=10cm]{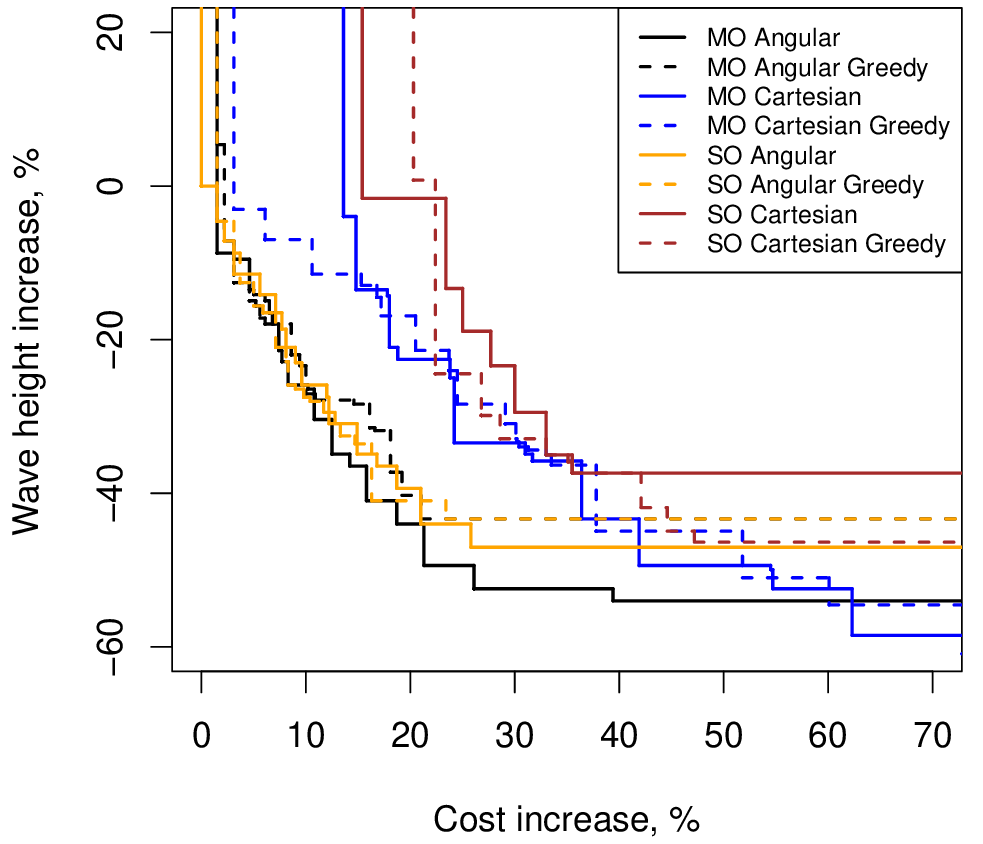}}
\caption{The comparison of Pareto fronts for the multi-objective algorithm with angular and Cartesian encodings. The solid lines are the basic algorithm, the dashed lines are the greedy modification}
\label{f:pareto_comp}
\end{figure}

The figure~\ref{f:pareto_comp} represents the comparison of Pareto fronts for multi-objective and single-objective algorithm. It can be seen, that the multi-objective algorithm with Angular encoding provides a better set of non-dominated solutions. Also, angular encoding allows increasing of effectiveness of all examined algorithms. The impact of the greedy heuristic is not so clear from this figure, but it provides competitive results.

\begin{figure*} 
\centerline{\includegraphics[width=16cm]{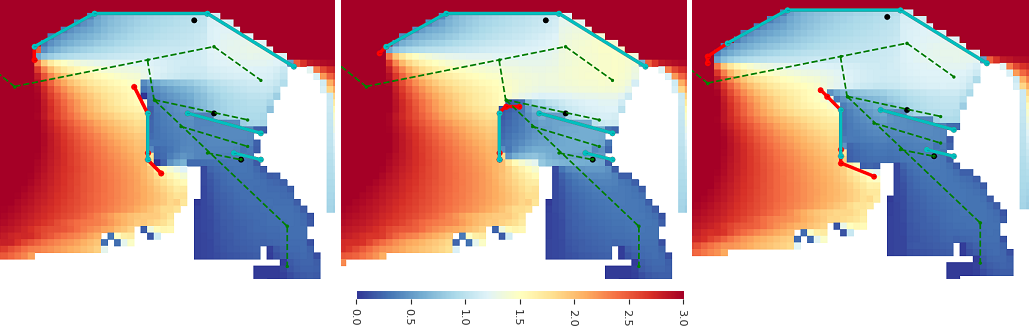}}
\caption{The variants of Pareto-optimal breakwaters' configurations and corresponding significant wave height fields (in meters). The coast is represented as a white mask}
\label{f:conf_res_example}
\end{figure*}

\begin{figure}
\centerline{\includegraphics[width=10cm]{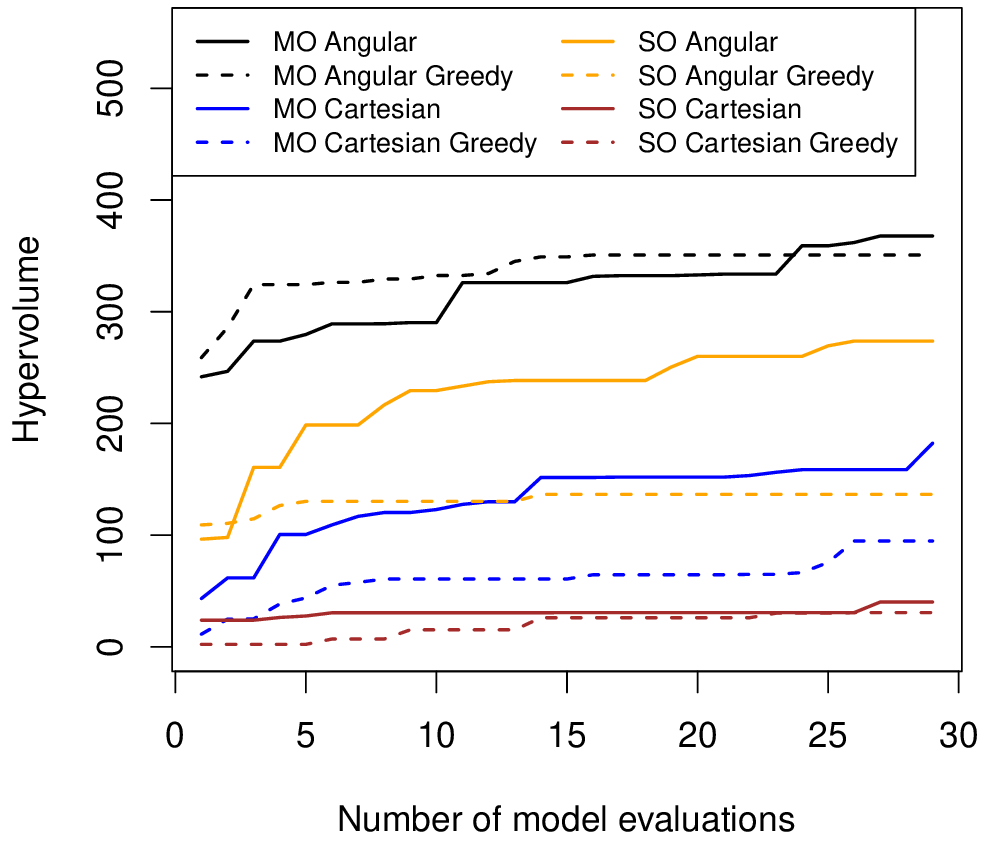}}
\caption{The convergence of the hypervolume metric in the full objectives space for the multi-objective algorithm with angular and Cartesian  encoding. The solid lines are the basic algorithm, the dashed lines are the greedy modification}
\label{f:hypervol}
\end{figure}

 The dependency of the hypervolume value from a number of wave model runs during evolution is presented in figure~\ref{f:hypervol} for different approaches. It can be noted that the multi-objective algorithm with angular encoding allows finding better solutions in comparison with other algorithms. Also, it's clear that the greedy heuristic included in the algorithm makes it possible to obtain a competitive solution at early generations for an evolutionary algorithm. Finally, all variants of the single-objective algorithm demonstrate worse performance.  

Several examples of simulations’ results obtained from the Pareto front are presented in figure~\ref{f:conf_res_example}. It can be seen that redundant segments are shrunk to zero length even without an additional structural complexity penalty. Since the optimisation approach is a Pareto-based, the breakwaters' configurations with different trade-offs between objective functions (mostly construction cost and wave height decrease) can be found it the final solutions set.

It should be noted that the choice of the most appropriate solution from the Pareto front is a complicated task \cite{horn1997f1}. The final decision should be based on an expert's experience, additional numerical and physical simulations and involvement of more detailed objectives.

\section{CONCLUSIONS}
\label{sec:conclusions}
In this paper, we formulate the task of breakwater layout optimal design identification as a computationally-intensive multi-objective optimisation problem. There are several evolutionary approaches validated using the Sochi port as a case study. The experiments were conducted with two different chromosome encodings (angular and Cartesian), two algorithms (single-objective differential evolution and multi-objective SPEA2). The multi-objective approach with angular chromosome encoding was selected as the most effective. Also, the effectiveness of greedy heuristics was evaluated as a part of the optimisation approach, and the faster convergence of the results set was obtained.

The source code of the implemented algorithms and software system used for experimental studies is available in the public repository (https://github.com/ITMO-NSS-team/breakwaters-evolutionary-optimisation).

The future extension of this research is planned as a generalisation of optimisation approaches to both attached and detached breakwaters, implementation of more realistic and detailed task cases, the involvement of additional synthetic and real-world test cases into experimental studies, and comparison of obtained configurations with expert solutions.

\label{sec:conl}

\section{Acknowledgements}

This research is financially supported by The Russian Scientific Foundation, Agreement \#19-11-00326.

The research is carried out using the equipment of the shared research facilities of HPC computing resources at Lomonosov Moscow State University.

\bibliographystyle{unsrt}  
\bibliography{references} 

\end{document}